\documentclass[graybox]{svmult}
\usepackage{type1cm}
\usepackage{makeidx}         
\usepackage{graphicx}        
\usepackage{multicol}
\usepackage[numbers]{natbib}
\usepackage[bottom]{footmisc}
\usepackage{newtxtext}
\usepackage[varvw]{newtxmath}
\usepackage[export]{adjustbox}

\begin{document}

\title*{Detecting Linguistic Diversity on Social Media}
\titlerunning{Detecting Linguistic Diversity on Social Media}
\author{Sidney Wong, Benjamin Adams, and Jonathan Dunn}
\institute{Sidney Wong \at Geospatial Research Institute,
\email{sidney.wong@pg.canterbury.ac.nz}
\and Benjamin Adams \at University of Canterbury,
\email{benjamin.adams@canterbury.ac.nz}
\and Jonathan Dunn \at University of Illinois Urbana-Champaign,
\email{jedunn@illinois.edu}
}
%
%
\maketitle
\abstract{
    This chapter explores the efficacy of using social media data to examine changing linguistic behaviour of a place. We focus our investigation on Aotearoa New Zealand where official statistics from the census is the only source of language use data. We use published census data as the ground truth and the social media sub-corpus from the Corpus of Global Language Use as our alternative data source. We use place as the common denominator between the two data sources. We identify the language conditions of each tweet in the social media data set and validated our results with two language identification models. We then compare levels of linguistic diversity at national, regional, and local geographies. The results suggest that social media language data has the possibility to provide a rich source of spatial and temporal insights on the linguistic profile of a place. We show that social media is sensitive to demographic and sociopolitical changes within a language and at low-level regional and local geographies.
}

\section{Introduction}
\label{sec:1}

    National censuses across the globe have been criticised for their ``high cost, low frequency, publication lag, limited geographic detail and limited breakdown of populations classifications into sub-categories'' \cite{gromme_2018}. Policy makers, non-government organisations, and researchers all use census data to determine the demographic profile of places and communities. In Aotearoa New Zealand (New Zealand), census information is the only source of data on languages for social scientists, linguists, and those working in language revitalisation, retention, and maintenance programmes \cite{bycroft_2016}. This reliance on a single data source is a risk leading us to consider alternative data sources and methodologies to collect this information. 
        
    In this chapter, we investigate the feasibility of using social media data as an alternative data source within the context of New Zealand. Based on current research, it is possible to observe linguistic behaviour of an underlying population using social media \cite{Eisenstein_2014}. Therefore, we use social media language data to examine the linguistic situation of New Zealand. We investigate the following research questions in our analysis:
    
    \begin{itemize}
      \item How do census data and social media data compare in terms of their basic characteristics and what they might tell us about language use variation over space and time? 
      \item And more importantly, how might we use social media data in place of official statistics when performing analyses based on language use information?
    \end{itemize}
    
    We analyse data from two sources: the New Zealand Census of Population and Dwellings (the census) and the web-based Corpus of Global Language Use (CGLU) \cite{dunn_2019}. We consider census as the ground truth on the linguistic situation of New Zealand and the social media language data from the CGLU is our alternative data source. We can compare the two data sets as they are both organised spatially. We use language identification models to identify the languages present in our social media language. We compare the efficacy of the two language identification models in our analysis. Following this data analysis step, we compute measures of linguistic diversity from our ground truth data and our alternative data source at national, regional, and local geographic levels. This allows us to understand the similarities and differences between census and social media language data.

\section{Background}
\label{sec:2}

\subsection{Linguistic Situation of New Zealand}
\label{subsec:1}

    Aoteaora is an island country located in the South Pacific Ocean. New Zealand has two official languages: te reo Māori and New Zealand Sign Language (NZSL). These languages accounts for 4.0\% and 0.5\% of languages used in New Zealand as of the 2018 Census. However, the most commonly spoken language in New Zealand is English which is the \textit{de facto} official language and accounts for 95.4\% of languages spoken in the population \cite{stats_nz_2020a}. The dominance of English across New Zealand society is a result of British colonisation and settlement beginning in the 19\textsuperscript{th} Century and subsequent impacts of globalisation \cite{kachru_1982}.
    
    Government preference for migrants from the United Kingdom and Australia and discriminatory legislation limiting migration from Asia has meant New Zealand society remained largely linguistically homogenous throughout the 19\textsuperscript{th} and 20\textsuperscript{th} Centuries. The demographic make-up of New Zealand significantly changed as a result of the Immigration Policy Review in 1986, which has allowed an increased number of migrants from non-Anglophone countries to immigrate to New Zealand \cite{henderson_2003}.
    
    This demographic change is evident in the most common languages used across New Zealand. The top ten most spoken languages as of the 2001 Census \cite{stats_nz_2023a} and the 2018 Census \cite{stats_nz_2020a} are presented in Table~\ref{tab:1}. While the top three most common languages (English, te reo Māori, and Samoan) remained the same, over half of the most common languages spoken in New Zealand originate from Asian countries such as Northern Chinese (including Mandarin), Hindi, Yue (Cantonese), Sinitic not further defined (which is used to signify an unspecified Sinitic (Chinese) language), and Tagalog. Respondents who did not provide a valid response or were too young to talk at the time of the census were excluded from Table~\ref{tab:1}.

    \begin{table}[!t]
     \renewcommand{\arraystretch}{1.3}
        \caption{Most common languages spoken in New Zealand in 2001 and 2018}
        \label{tab:1}
        \begin{tabular}{c l l}
            \hline\noalign{\smallskip}
            Rank & 2001 Census & 2018 Census\\
            \noalign{\smallskip}\svhline\noalign{\smallskip}
            1 & English & English\\
            2 & te reo Māori & te reo Māori\\
            3 & Samoan & Samoan\\
            4 & French & Northern Chinese\\
            5 & Yue & Hindi\\
            6 & German & French\\
            7 & NZSL & Yue\\
            8 & Northern Chinese & Sinitic not further defined\\
            9 & Dutch & Tagalog\\
            10 & Tongan & German\\
        \end{tabular}
    \end{table}

    We can see that in a period of less than twenty years, both the demographic and linguistic make-up of New Zealand have significantly changed. An increase in linguistic diversity coupled with the legislative recognition of te reo Māori in 1987 and NZSL in 2006 have increased the public consciousness to language rights and linguistic inclusion.

\subsection{Surveying Language}
\label{subsec:2}

    There are practical reasons in understanding the changing linguistic profile of New Zealand and allow communities to provide linguistically appropriate services across different populations. As of present, the most reliable source of language data for New Zealand comes from the census. The census provides the official count of all people and dwellings in New Zealand every five years \cite{statistics_new_zealand_2001}. This information is used to determine electoral boundaries and informs the distribution of public funding.
    
    The questions in the census are not static and have frequently changed over time. The changes depend on the needs of New Zealand's official statistics system. For example, questions relating to iwi and ethnic affiliation have only been included as part of the individual form since the 1991 Census. Other variables which were deemed no longer relevant such as race, ethnic origin, or nationality have since been removed. The language census topic was only included as part of the 1996 Census.
    
    The \textit{languages spoken} variable is derived from the language census topic question in the individual form. More specifically, the questions asks ``In which language(s) could you have a conversation about a lot of everyday things?''. This question is a multiple response variable which means participants can include up to six languages which are coded according to the Language Standard Classification 1999 \cite{stats_nz_2021b}. In a multiple response variable, each response is counted towards each applicable language classification. Other modalities of language such as signed languages (e.g., NZSL, American Sign Language) are also included as part of the question.
    
    The languages spoken variable is a priority three variable which means the variable does not directly fit with the main purpose of the census, but this information is still important to certain populations and communities \cite{stats_nz_2021b}. The language census topic and the languages spoken variable serves the following functions:
    
     \begin{enumerate}
        \item To formulate, target and monitor policies and programmes to revitalise the Māori language as an official language of New Zealand.
        \item As an indicator of iwi vitality and cultural resources.
        \item To assess the need to provide multi-lingual pamphlets and other translation services in a variety of areas such as education, health and welfare.
        \item To evaluate and monitor existing language education programmes and services.
        \item To provide information for television and radio programmes and services.
        \item To understand the diversity and diversification of the New Zealand population over time, as well as language maintenance, retention and distribution.
     \end{enumerate}
    
    One limitation of the language census topic is that it does not collect written linguistic ability or fluency. It is not possible to determine linguistic competence across different modalities from the language census topic alone. Only Te Kupenga, a sample survey of 8,500 Māori aged 15 years and over living across New Zealand, includes questions regarding written ability and fluency in te reo Māori \cite{stats_nz_2018}. This means the census may not meet the needs of heritage language revitalisation efforts for other culturally and linguistically diverse communities.
    
    Despite the benefits of the census, the national survey is a costly endeavour. The 2023 Census was estimated to cost NZ\$250 million - almost double the cost of the 2018 Census \cite{williams_2022}. The success of a census, measured by its response rate, is susceptible to social, environment, and political factors. The proposed 2011 Census was postponed to 2013 due to the 2010-2011 Canterbury Earthquakes. There is also a trend in declining response rates to surveys \cite{greaves_2020}. This downward trend was reflected in the results of the 2018 Census which was known for its low response rate of 83.3\% \cite{stats_nz_2019}. This is significant 8.9\% decline from the 2013 Census.
    
    In the case of the language census topic, only 83.8\% of responses came from the 2018 Census, 8.2\% of responses were derived from 2013 Census data, and the remaining 8.0\% of responses were derived through statistical imputation \cite{stats_nz_2021b}. Despite the low response, the data quality of the languages spoken variable was rated high quality. This is particularly concerning for policy makers and researchers who use this information to determine the success of language revitalisation programmes.

\subsection{Alternative Data Sources}
\label{subsec:3}

    The five-year (and in some cases ten-year for some countries) census cycles which have traditionally met the requirements to simply count a population. This does not meet the needs of data hungry public and private sector organisations who require more contemporaneous demographic insights on populations. If the purpose of a census is to understand the changing demographic profile of a population, then perhaps there are other data sources and methods to meet this need.
    
    One suggested method is the use administrative data as an alternative to a conventional census survey \cite{obyrne_2014}. Administrative data includes all the transactional data held by central government ministries and departments. This information is currently managed by Stats NZ as part of the Integrated Data Infrastructure (IDI) \cite{milne_2019}. For example, information about individuals in New Zealand such as location and ethnicity data are held by the Ministry of Health, education and training data are held by the Ministry of Education, work and income data are held by Inland Revenue to list a few. This method will only be suitable for some topics and not others.
    
    There are currently no administrative sources identified which is suitable to replace the languages spoken variable \cite{bycroft_2016}. The only alternative sources for language information from other official sources are the biennial General Social Survey (GSS), a sample survey of 8,000 individuals in households across New Zealand aged 15 years and over \cite{stats_nz_2023b}, and enrolment data from the Ministry of Education \cite{moe_2023}. This means we are unlikely to get good quality data about language equivalent to a census from administrative sources \cite{obyrne_2014}.
    
    These data gaps coupled with technical issues surrounding sensitive demographic attributes such as ethnicity, the lack of consent sought from individuals, and barriers in access to the data in the IDI raise legislative and ethical issues as a source of official statistics. These issues have serious consequences for the government and its commitment to Te Tiriti o Waitangi in upholding Māori Data Sovereignty (MDS) \cite{greaves_2023}. In essence, MDS ensures Māori communities have the ability to exercise power over usage and outputs of data produced by Māori and about Māori \cite{tmr_2018}.
    
    Beyond administrative data, social media has been offered as an alternative data source for national censuses and surveys outside of New Zealand \cite{gromme_2018}. Early work in the United Kingdom investigating the feasibility of using social media data to derive demographic information found that gender and language information from Twitter, also known as X, is proportional to results observed in the 2011 Census \cite{sloan_2013}.
    
    There were an estimated 4.99 million internet users in New Zealand in January 2023. This is equivalent to an internet penetration rate of 95.9\% out of New Zealand's estimated population of 5.21 million \cite{Kemp_2023}. More specifically, there was an estimated 4.24 million social media users in New Zealand. This means over 80.0\% of New Zealand's population have subscribed to a social media platform. Based on this level of coverage, social media may be a feasible alternative data source. The user base for different social media platforms have been included in Table~\ref{tab:2}.
    
    \begin{table}[!t]
        \caption{Social media platform users in New Zealand}
        \label{tab:2}
        \begin{tabular}{l c}
            \hline\noalign{\smallskip}
            Platform & Users (in millions)\\
            \noalign{\smallskip}\svhline\noalign{\smallskip}
            Facebook & 2.95\\
            Youtube & 4.24\\
            Instagram & 2.15\\
            TikTok & 1.65\\
            LinkedIn & 2.50\\
            Snapchat & 1.45\\
            X (Twitter) & 0.79\\
            Pinterst & 0.67\\
        \end{tabular}
    \end{table}

\subsection{Digital Language and Place}
\label{subsec:5}

    Of the different social media platforms, X (Twitter) is by far the greatest contributor of digital language data despite representing 15.2\% of New Zealand. A benefit of X (Twitter) is the massive volume of publicly available data. X (Twitter) provides researchers access to their data including tweet and user information through their Application Programming Interface (API). Researchers with `Essential' access can retrieve up to 500 thousand tweets per month while those who qualify for `Academic Research' access can retrieve up to 10 million tweets per month. 
    
    Researchers can also access temporal and spatial information for each tweet through the API based on X (Twitter)'s user-enabled geotagging feature \cite{marti_2019}. The search polygon shape is determined by the type of data request: rectangular for real-time tweets and circular for historical tweets. There are no limits to the search polygon size, but there is no guarantee all tweets within a geographic area will be retrieved.
    
    There are some limitations to this data. The number of geotagged tweets only account for a small proportion of tweets as the geotagging feature is disabled by default. Sloan et al. \cite{sloan_2013} found that only 0.85\% of tweets were geotagged in a sample of 113 million tweets. There are also significant demographic differences between the type of users who enable geotagging based on the user's perceived age, social class, language background, and user interface language \cite{sloan_2015}.
    
    X (Twitter)'s user base may not be a balanced sample of an underlying population. In the US context, X (Twitter) users tend to skew towards younger urban users who come from an ethnic minority background \cite{mislove_2011}. We must consider the types of biases inherent to digital language and place data when making claims about linguistic behaviour on social media.
    
    Recent work has shown progress in addressing these limitations. Dunn et al. \cite{dunn_2020a} found that digital language behaviour was sensitive to real-world events. The study tracked changes in national measures of linguistic diversity over the course of the COVID-19 pandemic. They were able to account for non-local bias as nationwide lockdowns limited travel and migration internationally and domestically.

\section{Data and Methods}
\label{sec:3}

    \begin{figure}[!t]
        \sidecaption
        \includegraphics[width=\textwidth]{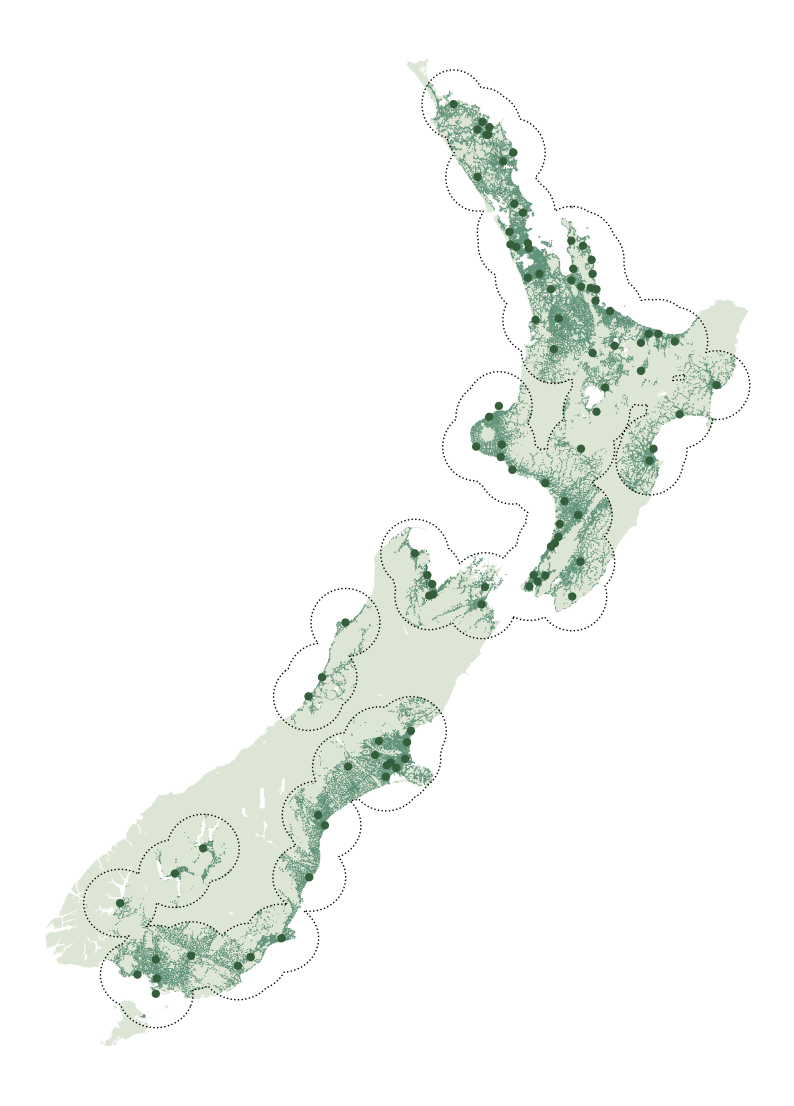}
        \caption{Settlements within CGLU catchment area.}
        \label{fig:inside_catchment}
    \end{figure}
    
    \begin{figure}[!t]
        \sidecaption
        \includegraphics[width=\textwidth]{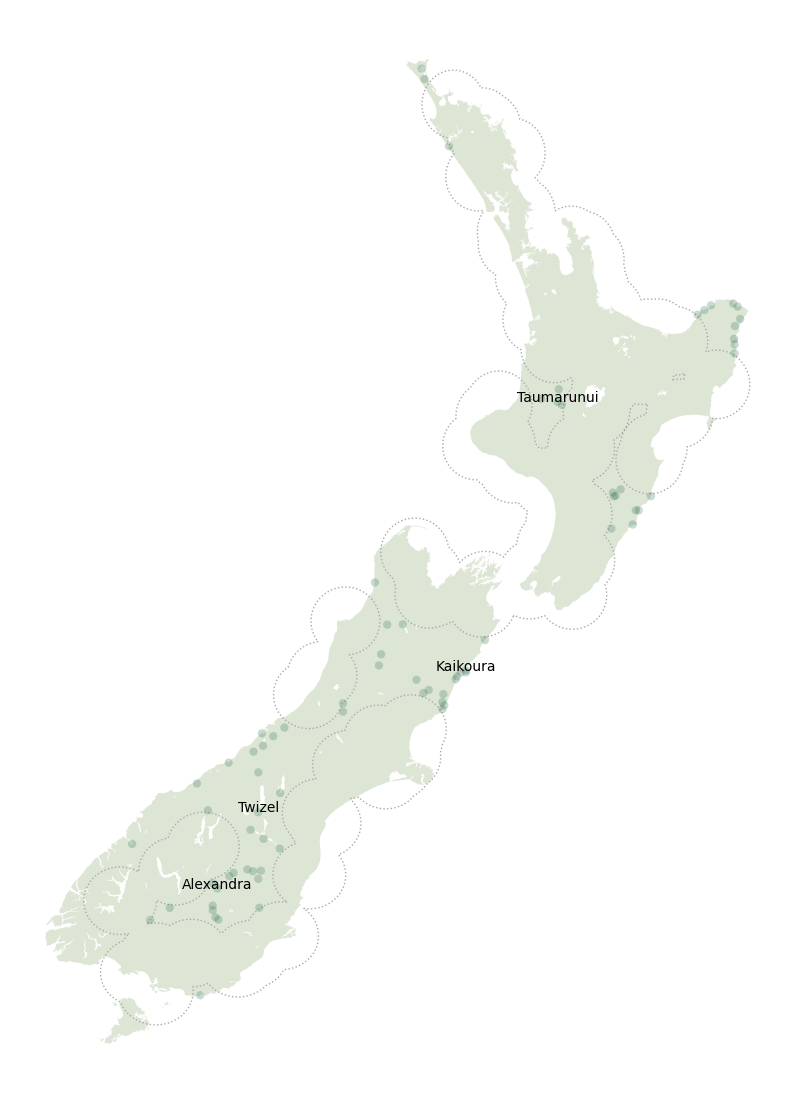}
        \caption{Settlements outside CGLU catchment area.}
        \label{fig:outside_catchment}
    \end{figure}

\subsection{Corpus of Global Language Use}
\label{subsec:8}

    The Corpus of Global Language Use (CGLU) is a web-based corpus which consists of the web sub-corpus retrieved through Common Crawl and the social media sub-corpus retrieved through the X (Twitter) API \cite{dunn_2019}. Data collection has been on-going since 2017. Our focus is the social media sub-corpus. 
    
    The sub-corpus is coded with both temporal and spatial information. Each tweet has been coded for broad geographic region, country of origin, nearest city, date, a corresponding fifteen-character geohash , and the content of the tweet itself. The geohash is derived from the latitude and longitude information linked to geotagged enabled tweets. The data collection points come from the global Geonames data set \cite{geonames_2023}. We filtered the social media sub-corpus for tweets originating from within New Zealand. There are one hundred data collection points across New Zealand. The social media sub-corpus contains geotagged enabled tweets within a 50-kilometre radius for each data collection point. The data set does not contain duplicates.
       
    We linked each of the data collection points to one of the sixteen regional council areas. We visualised the data collection points in Figure~\ref{fig:inside_catchment}. A majority of the data collection points are in Te Ika a Māui (the North Island). This is expected as more than a third of the population resides in Te Ika a Māui. Te Waipounamu (the South Island) is sparsely populated despite being the larger of the two islands. The data catchment area accounts for 97.6\% of the estimated resident population as of December 2022. Inversely, Figure \ref{fig:outside_catchment} visualises all small urban areas (light green) situated outside the data catchment area. This area amounts for a estimated population of 56,904 as of December 2022 \cite{stats_nz_2023b}. Small urban areas and rural settlements with an estimated population greater than 1,000 people as of 2022 includes: Taumarunui (est. 4,830), Kaikōura (est. 2,330), Twizel (est. 1,780), and Alexandra (est. 6,010). Overall, the \textsc{cglu} has good coverage of New Zealand despite missing data from these small urban settlements
    
    More detailed information on the individual data collection points are presented in Table~\ref{tab:5} the Appendix. We do not have data linked to the Nelson region. With reference to Figure~\ref{fig:inside_catchment}, tweets originating from the Nelson region are captured by data collection points located in the neighbouring Tasman and Marlborough regions.

\subsection{New Zealand Census of Population and Dwellings}
\label{subsec:9}

    We use published census data as the ground truth data set. It is a suitable source to determine the ground truth linguistic situation of New Zealand as the census is a national count of all individuals in New Zealand. Published census data is publicly accessible online. This data is aggregated and confidentialised. We retrieved the national-level and regional-level data from NZ.Stat \cite{stats_nz_2020a}. The data set includes data from the 2006, 2013, and 2018 census cycles. As we only have access to social media data from 2017 on wards, the focus of our study is on the 2018 Census.
    
    We removed additional metadata information as part of the data cleaning process. Signed languages such as NZSL were removed from the analytical data set. These were excluded from the current analysis as they do not have a corresponding written orthography identifiable through the language identification models. Furthermore, unspecified languages from the statistical classification in 'Other' and ‘None (e.g. too young to talk)’ were also removed from the analytical data set.
     
    We retrieved additional population information for New Zealand from NZ.Stat \cite{stats_nz_2020b} and geographic information from Datafinder \cite{stats_nz_2021a}. A demographic summary of New Zealand and the regional council areas are available in Table~\ref{tab:6} of the Appendix.

\subsection{Language Identification}
\label{subsec:10}

    \begin{figure}[!t]
        \sidecaption
        \includegraphics[width=\textwidth]{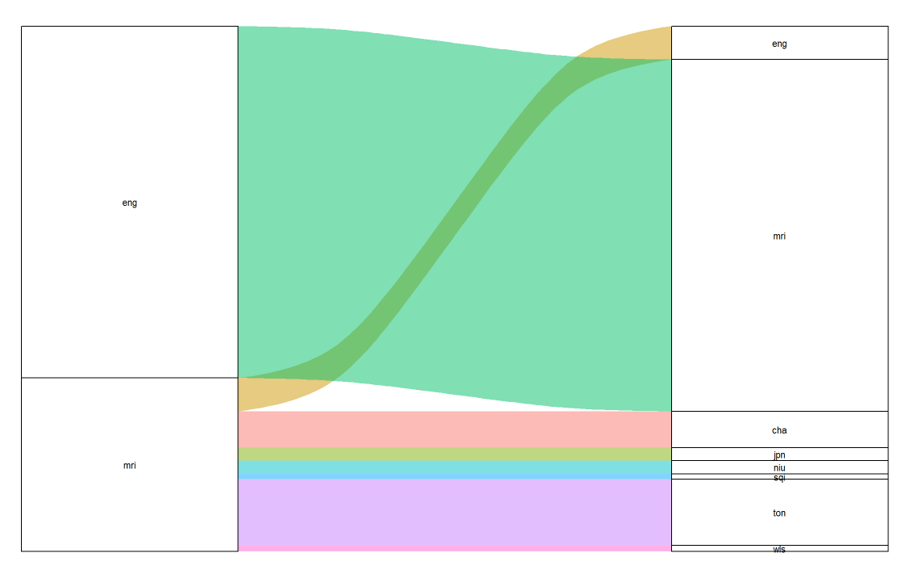}
        \caption{Differences between language identification models idNet (\textit{left}) and pacificID (\textit{right}).}
        \label{fig:error_analysis}
    \end{figure}
    
    The first data processing step in our study is to identify the predominate language for each tweet. Unlike the census language topic which is based on self-rated ability, the language condition for each tweet is not implicitly available from the CGLU. Although X (Twitter) provides support for 34 written languages on sign-up, 33.0\% of tweets were in a language that was different from the user interface in a sample of 113 million tweets \cite{sloan_2013}. 
    
    We use the idNet language identification classification model to automatically code the primary language of a tweet \cite{dunn_2020b}. The package has a high accuracy with a reported F1-score above 0.95 for 464 languages based on 50-character language samples. The second language classifier we have used for this study is the pacificLID package \cite{dunn_2022}. This language classifier was especially adapted to identify Austronesian languages (e.g., te reo Māori). This is particularly useful in an New Zealand context. We visualised the classification differences between the language identification models in Figure \ref{fig:error_analysis}.  Both classifiers were included as part of our analysis for comparability as there may be classification errors (i.e., as a result of code switching). 
    
\subsection{Linguistic Diversity}
\label{subsec:11}

    \subsection{Language Identification}
    \label{subsec:12}
    \begin{table}[!t]
        \caption{Most common languages in 2018}
        \label{tab:3}
        \begin{tabular}{l l l l}
            \hline\noalign{\smallskip}
            Rank & 2018 Census & idNet & pacificLID\\
            \noalign{\smallskip}\svhline\noalign{\smallskip}
            1 &  English & English & English\\
            2 &  te reo Māori & Portuguese & Portuguese\\
            3 &  Samoan & Japanese & Thai\\
            4 &  Northern Chinese & Tagalog & Japanese\\
            5 &  Hindi & Spanish & Spanish\\
            6 &  French & Indonesian & Tagalog\\
            7 &  Yue & Arabic & Malaysian\\
            8 &  Sinitic not further defined & French & French\\
            9 &  Tagalog & Korean & Arabic\\
            10 &  German & Thai & Korean\\
        \end{tabular}
    \end{table}

    The last data processing step is to calculate measure of linguistic diversity for the various levels of geography (national, regional, and local). A simple implementation of this measure is the concentration ratio (CR) which we can use as a proxy for linguistic diversity \cite{hirschman_1945}. A CR is used to determine the market structure and competitiveness of the market and provides a range between 0\% to 100\%. Common CR measures include 4-firm (CR$_{\text{4}}$) and 8-firm (CR$_{\text{8}}$). We used a 10-firm CR ($CR$$_{\text{10}}$). This means for each geographic location, we selected the top ten most common languages in use. The CR is calculated as in Equation~\ref{eq:1}.
    
    \begin{equation}
        CR_n = C_1 + C_2 + ... + C_n
        \label{eq:1}
    \end{equation}
    
    Where: \textit{C$_{\text{n}}$} defines the share of the $n\textsuperscript{th}$ largest languages as a \% of a population and \textit{n} defines the number of languages included in the CR calculation.
    
    A $CR$$_{\text{10}}$ measure under 0.40 suggests low concentration (i.e., perfect competition). A $CR$$_{\text{10}}$ measure between 0.40 to 0.70 suggests medium concentration (i.e., an oligopoly). A $CR$$_{\text{10}}$ measure over 0.70 suggests high concentration (i.e., a monopoly). This means the the lower the measure, the higher the levels of linguistic diversity.
        
\section{Results}
\label{sec:4}
    
    \begin{figure}
        \sidecaption
        \includegraphics[width=\textwidth]{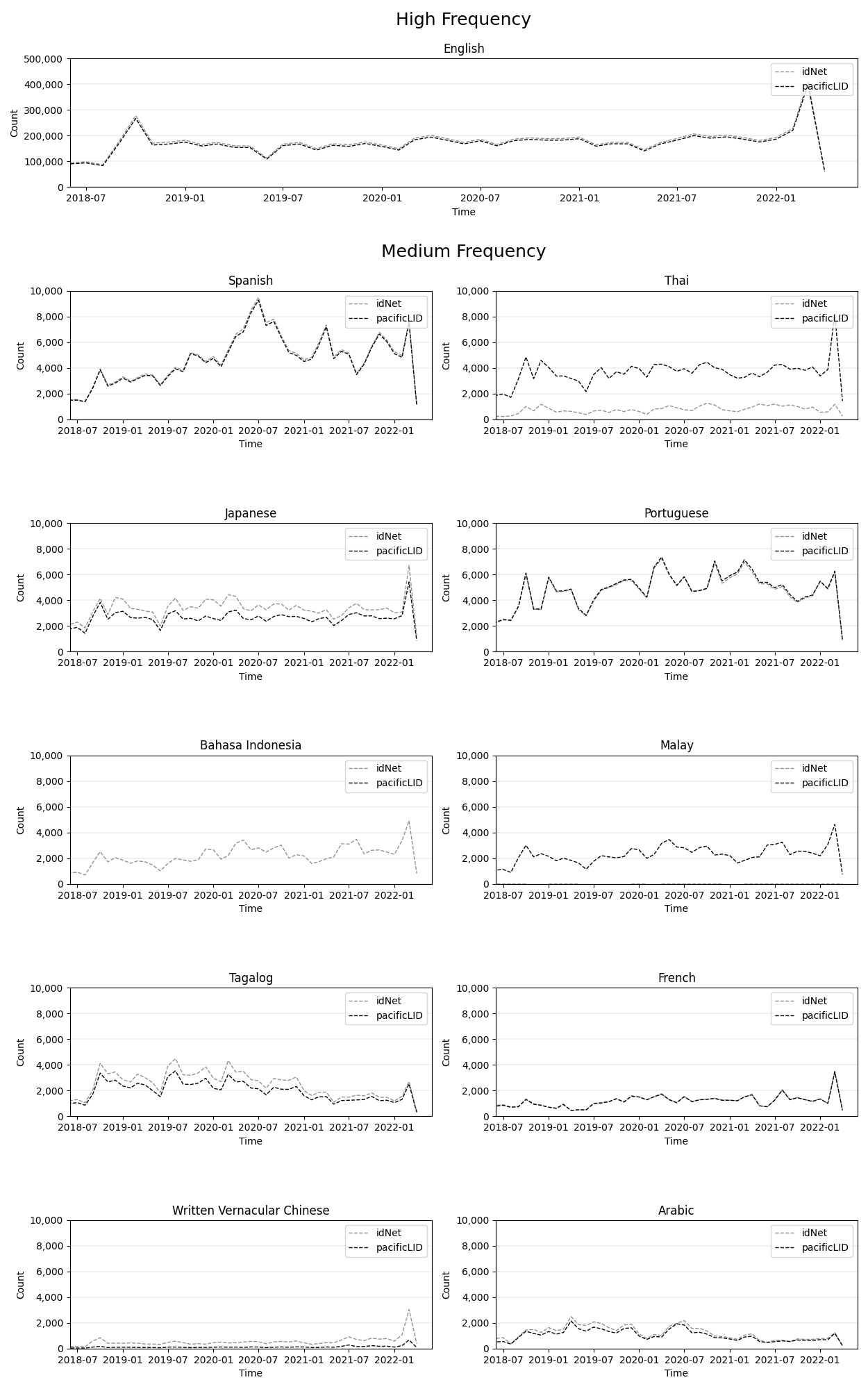}
        \caption{High and Medium Frequency Languages on X (Twitter)}
        \label{fig:2}
    \end{figure} 
    
    \begin{figure}
        \sidecaption
        \includegraphics[width=\textwidth]{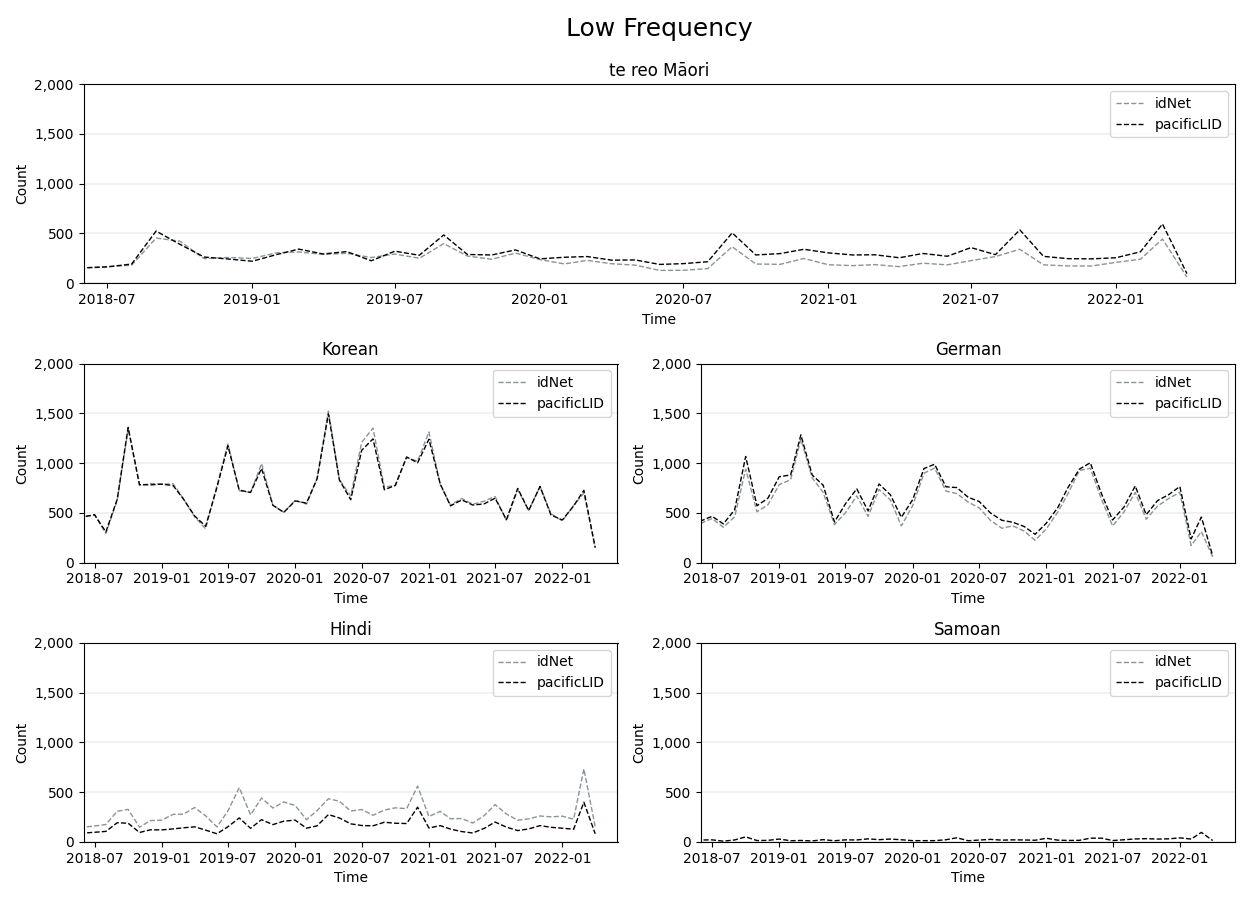}
        \caption{Low Frequency Languages on X (Twitter)}
        \label{fig:3}
    \end{figure} 
    
    We observed a high level of agreement between the idNet and the pacificLID models. There were 76,007 mismatches between the two language identification models. The rate of mismatch between the two packages was equivalent to 0.76\%. A closer inspection of mismatched tweets in English and te reo Māori found that a majority of tweets were reclassified from English to te reo Māori between idNet and pacificLID. 
    
    In the case of te reo Māori tweets identified with idNet, most tweets were reclassified to Tongan and the remaining tweets were equally reclassified to English, Chamorro, Japanese, Niuean, Albanian, and Wallisian. The reclassification of tweets in te reo Māori to Chamorro, Niuean, Tongan, and Wallisian was expected as the pacificLID package was trained to be more sensitive to Austronesian languages. 
    
    The reclassified tweets could be corrections, or indeed classification errors as some tweets which were predominately in English were reclassified as te reo Māori. Other reasons for the mismatches between the idNet and the pacificLID could be the result of code-switching or other translanguaging practices that were not captured by a classification type language identification model. 
    
    We identified 403 distinct languages within the sub-corpus. This number is significantly higher than the 196 languages listed in the Language Standard Classification 1999. The nature of written language differs significantly from spoken language. There is not always a one-to-one relationship between the two modes of language. In some cases there could be a zero-to-one, one-to-many, or many-to-many relationships between modes. This means language varieties represented in the census data may not appear in the social media sub-corpus and vice versa. 
    
    Table~\ref{tab:3} compares the top ten most common languages from the latest census and the social media language data. English was the most common language across the two models. This was expected. However, there were a few unexpected results. For example, te reo Māori, Samoan, Hindi, German, or any of the Sinitic languages were rarely observed on X (Twitter). While census data only provides spatial granularity to neighbourhood level geographies, social media data provides both spatial and temporal granularity. This is because census cycles occur once every five (or ten years), while social media platforms can capture data contemporaneously. Figure~\ref{fig:2} and Figure~\ref{fig:3} demonstrate this by visualising the monthly frequency count of tweets in some language conditions.
    
    Figure~\ref{fig:2} includes languages with high (where \textit{y}-limit is 500,000) and medium (where \textit{y}-limit is 10,000) frequency counts while Figure~\ref{fig:3} includes languages with low (where \textit{y}-limit is 10,000). The dashed lines on the figure compares the results from idNet (in grey) and pacificLID (in black). We can see a high level of consistency between the two language identification models for English, Spanish, Portuguese, and French in Figure~\ref{fig:2} and Korean and German in Figure~\ref{fig:3}. 
    
    It is clear that the level of consistency is not the result of data availability (as shown by the monthly frequency count on the \textit{y}-axis), but the efficacy of the language identification model itself. There are some severe irregularities between the two models particularly in Japanese, Tagalog, and written vernacular Chinese, Tagalog in Figure~\ref{fig:2} and te reo Māori and Hindi in Figure~\ref{fig:3}. The discrepancies between Bahasa Indonesia and Malay is a result of how these languages are coded in the two models. These differences require a deeper analysis beyond the scope of this chapter.
    
    Despite these irregularities, there is value in this information as we can conduct time series analyses on different language conditions. For te reo Māori in particular as shown in Figure~\ref{fig:3}, we can observe a seasonal peak of tweets during the second half of the year. This increase in frequency coincides with Mahuru Māori where participants are encouraged to use te reo Māori in all facets of everyday life. This suggests we can observe real-time changes to the linguistic profile of New Zealand based on significant cultural movements.

\subsection{Linguistic Diversity}
\label{subsec:13}

    \begin{figure}[!t]
        \sidecaption
        \includegraphics[width=\textwidth]{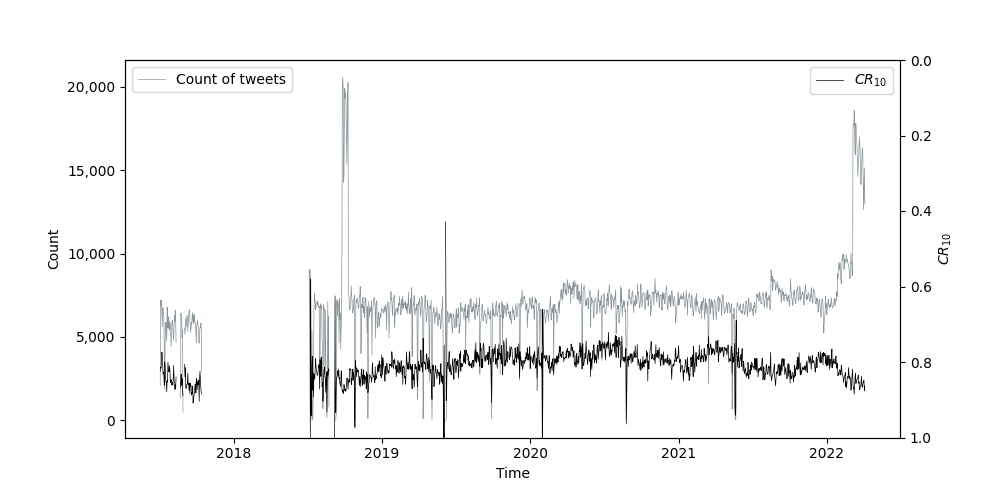}
        \caption{Monthly frequency counts of tweets and corresponding $CR$$_{\text{10}}$ measures}
        \label{fig:4}
    \end{figure}
    
    Now that we have confirmed that the language identification models can suitably identify languages in social media sub-corpus, we can calculate measures of linguistic diversity based on the 10-fold concentration ratio ($CR$$_{\text{10}}$) measure. The initial measure of the national $CR$$_{\text{10}}$ measure for 2006 was 0.76, 0.81 in 2013, and 0.79 in 2018. These measures from the census suggest that New Zealand is typically linguistically homogeneous. The $CR$$_{\text{10}}$ measures from the social media sub-corpus for 2018 was 0.79 (idNet) and 0.72 (pacificLID). We can see from this national measure, idNet was more consistent with the 2018 Census. Once again, these results suggest that even the digital presence of New Zealand is typically linguistically homogeneous.
    
    These national-level measures provide little detail on how the linguistic profile of New Zealand has changed over time. We are again interested in the temporal granularity of the social media language data and how the frequency counts of tweets over time (i.e., the sampling) may have an effect on linguistic diversity. We can analyse the stability of the time series visually. Figure~\ref{fig:4} is a multiple line graph with two \textit{y}-axes. The primary \textit{y}-axis represents the daily counts of tweets (in grey) and the secondary \textit{y}-axis represents the linguistic diversity (in black). We have only included tweets where idNet and pacificLID matched. We inverted the secondary \textit{y}-axis to improve interpretability.
    
    The most striking result is the missing data between 2017-2018 and the spikes of frequency counts. These are clearly outliers in the data collection. When we discount these outliers, we can observe a stable relationship between the frequency of tweets and the $CR$$_{\text{10}}$ which suggests the time series is stationary. A stationary time series is particularly important if we were to model trends from the social media language data.

\subsection{Regional Insights}
\label{subsec:14}

    \begin{table}[!t]
        \caption{$CR$$_{\text{10}}$ measures for 2018 by regional council areas}
        \label{tab:4}
        \begin{tabular}{l c c c}
            \hline\noalign{\smallskip}
            Region Name & Census & idNet & pacificLID\\
            \noalign{\smallskip}\svhline\noalign{\smallskip}
            Northland & 0.76 & 0.56 & 0.52\\
            Auckland & 0.60 & 0.79 & 0.73\\
            Waikato & 0.76 & 0.81 & 0.75\\
            Bay of Plenty & 0.75 & 0.80 & 0.73\\
            Gisborne & 0.70 & 0.88 & 0.48\\
            Hawkes Bay & 0.78 & 0.95 & 0.86\\
            Taranaki & 0.85 & 0.49 & 0.44\\
            Manawatū-Wanganui & 0.80 & 0.71 & 0.66\\
            Wellington & 0.72 & 0.89 & 0.82\\
            West Coast & 0.89 & 0.66 & 0.60\\
            Canterbury & 0.80 & 0.79 & 0.72\\
            Otago & 0.82 & 0.90 & 0.83\\
            Southland & 0.87 & 0.89 & 0.81\\
            Tasman & 0.86 & 0.58 & 0.54\\
            Marlborough & 0.85 & 0.88 & 0.78\\
        \end{tabular}
    \end{table}
    
    Of interest to policy makers and researchers is the potential to use alternative data sources to provide insights at regional and local geographies. Table~\ref{tab:4} provides the $CR$$_{\text{10}}$ measures for the census and social media sub-corpus for 2018 year by regional council areas. Once again, we compare the results from idNet and pacificLID.
     
    Firstly, the $CR$$_{\text{10}}$ measures from the census differ between regional council areas. Auckland is the most linguistically diverse, while the West Coast is the least linguistically diverse. Regions in Te Ika a Māui (with the exception of Taranaki) are more linguistically diverse than regions in Te Waipounamu. As we compare the $CR$$_{\text{10}}$ measures from the different data sources we see greater variability. Surprisingly, Taranaki region is the most linguistically diverse according to the social media sub-corpus which significantly contrasts its corresponding measure from the census. There is little consistency between the census measures and the social media language measures with the exception of Canterbury, Southland, and Marlborough regions which have similar levels of linguistic diversity.
    
    \begin{figure}
        \sidecaption
        \includegraphics[width=\textwidth]{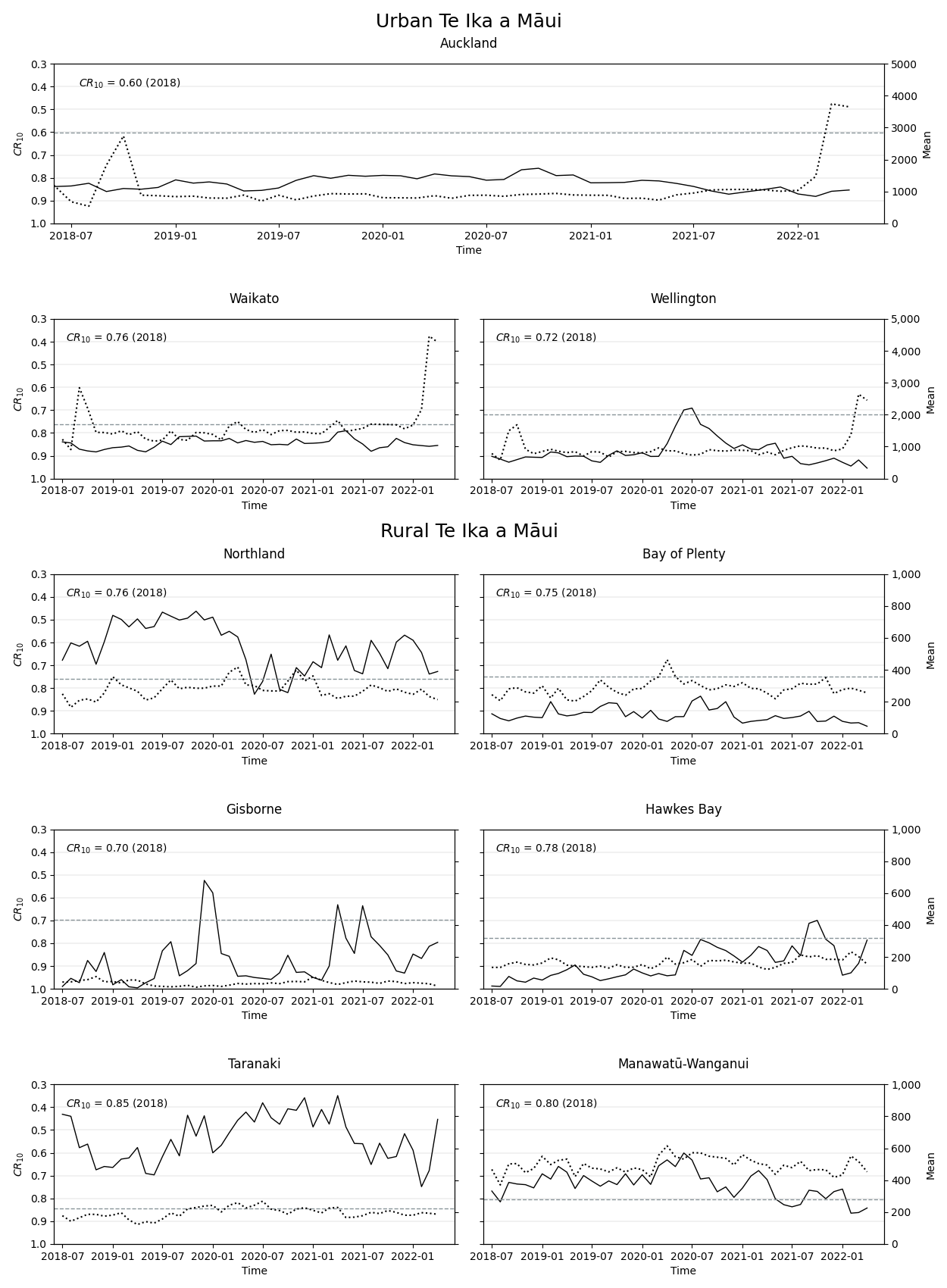}
        \caption{Monthly $CR$$_{\text{10}}$ Measures for Te Ika a Māui by regional council area}
        \label{fig:5}
    \end{figure}
    
    \begin{figure}
        \sidecaption
        \includegraphics[width=\textwidth]{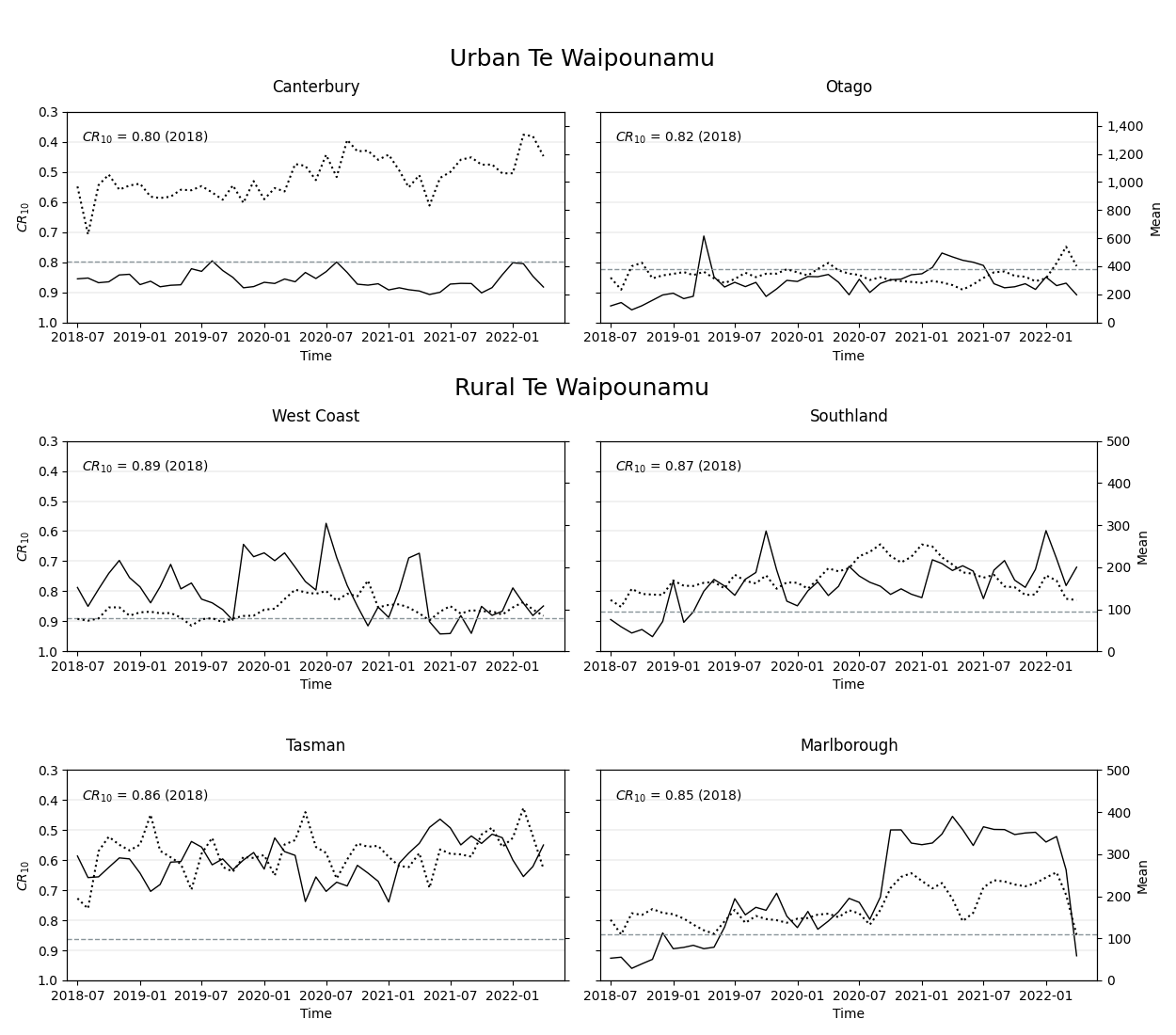}
        \caption{Monthly $CR$$_{\text{10}}$ Measures for Te Waipounamu by regional council area}
        \label{fig:6}
    \end{figure}
    
    Another striking observation is that the $CR$$_{\text{10}}$ measures derived from the pacificLID exhibited greater levels of linguistic diversity than census or idNet derived measures (with the exception of Auckland). The fact that we can only observe these differences at the regional level suggest a downstream effect of the 76,007 mismatches from the two language identification models. A regional breakdown of the number of tweets and proportion of tweets is in Table~\ref{tab:5} in the Appendix.
    
    We carried out a simple non-parametric test between the $CR$$_{\text{10}}$ observed in Table~\ref{tab:4} and demographic information derived from census and X (Twitter) as presented in Table~\ref{tab:5} and Table~\ref{tab:6} of the Appendix. We did not observe a relationship between the census derived $CR$$_{\text{10}}$ measures and X (Twitter) derived $CR$$_{\text{10}}$ measures at the regional level. The correlation coefficient (Spearman's Rho) between the census and idNet was -0.27 and pacificLID was -0.08. We observed a moderate negative relationship between the census derived $CR$$_{\text{10}}$ measure with population density (-0.73**) and a weak positive relationship with median age (0.56*).
    
    We did observe a statistically significant strong positive correlation coefficient between the $CR$$_{\text{10}}$ measures derived from idNet and pacificLID which was 0.80***. This suggests a high level of consistency between the language identification models. There was no relationship between X (Twitter) derived $CR$$_{\text{10}}$ measures and the demographic information in Table~\ref{tab:6}. However, we observed a weak negative relationship between census derived $CR$$_{\text{10}}$ with the proportion of tweets per region in relation to corpus size (-0.51*).
    
    The results from the Spearman's Rho suggest there is a weak association between demographic measures derived from census and social media sub-corpus at regional geographies. It is possible the small sample of regional geographies is not sufficient to identify a consistent relationship between the two sources of data.
    
    Figure~\ref{fig:5} and Figure~\ref{fig:6} provide a monthly breakdown of the $CR$$_{\text{10}}$ by regional council area over time on the primary \textit{y}-axis (solid in black). The regions are grouped by islands and urban-rural in order to standardise the \textit{y}-axis limit. We inverted the \textit{y}-axis to improve interpretability. We have only included the $CR$$_{\text{10}}$ measures derived from the pacificLID. The monthly mean frequency count of tweets are shown on the secondary \textit{y}-axis (dotted in black). Furthermore, We included 2018 Census $CR$$_{\text{10}}$ measures as our baseline to see how linguistic diversity on social media compares with the ground truth measures of linguistic diversity for each regional council area (dotted in grey).
    
    We can see from Figure~\ref{fig:5} and Figure~\ref{fig:6} that a higher monthly mean frequency count corresponds to a more stable $CR$$_{\text{10}}$ measure. This is shown in the urban regional councils such as Auckland, Waikato, and Wellington regions in Te Ika a Māui as shown in Figure~\ref{fig:5} and Canterbury and Otago regions in Te Waipounamu as shown in Figure~\ref{fig:5}. Furthermore, linguistic diversity from the social media sub-corpus is consistently lower for all urban regions than the census, while the opposite is true for rural regions (with the exception of Bay of Plenty, Gisborne, and Hawke's Bay region.

\subsection{Case Study: Wellington}
\label{subsec:15}

    \begin{figure}[!t]
        \sidecaption
        \includegraphics[width=\textwidth]{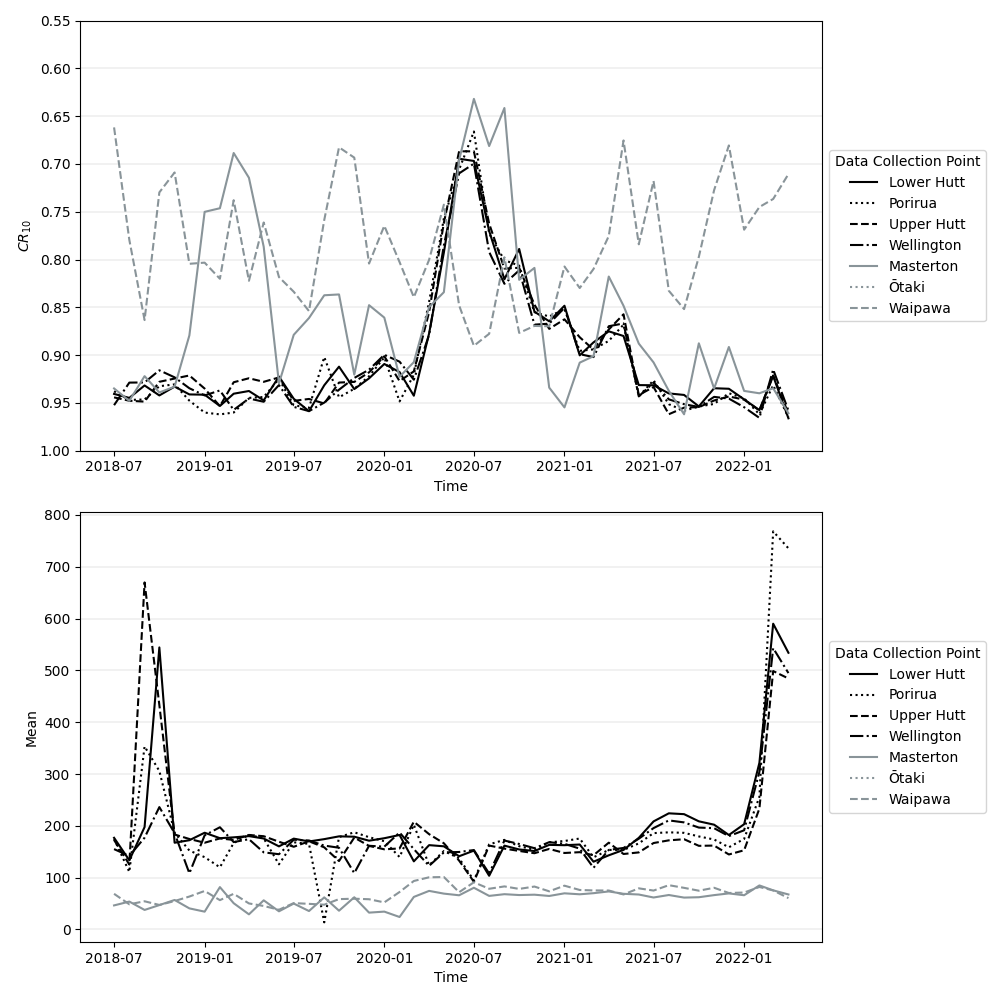}
        \caption{Monthly $CR$$_{\text{10}}$ measures and mean counts for data collection points in the Wellington region}
        \label{fig:8}
    \end{figure}
    
    \begin{figure}[!t]
        \sidecaption
        \includegraphics[width=\textwidth]{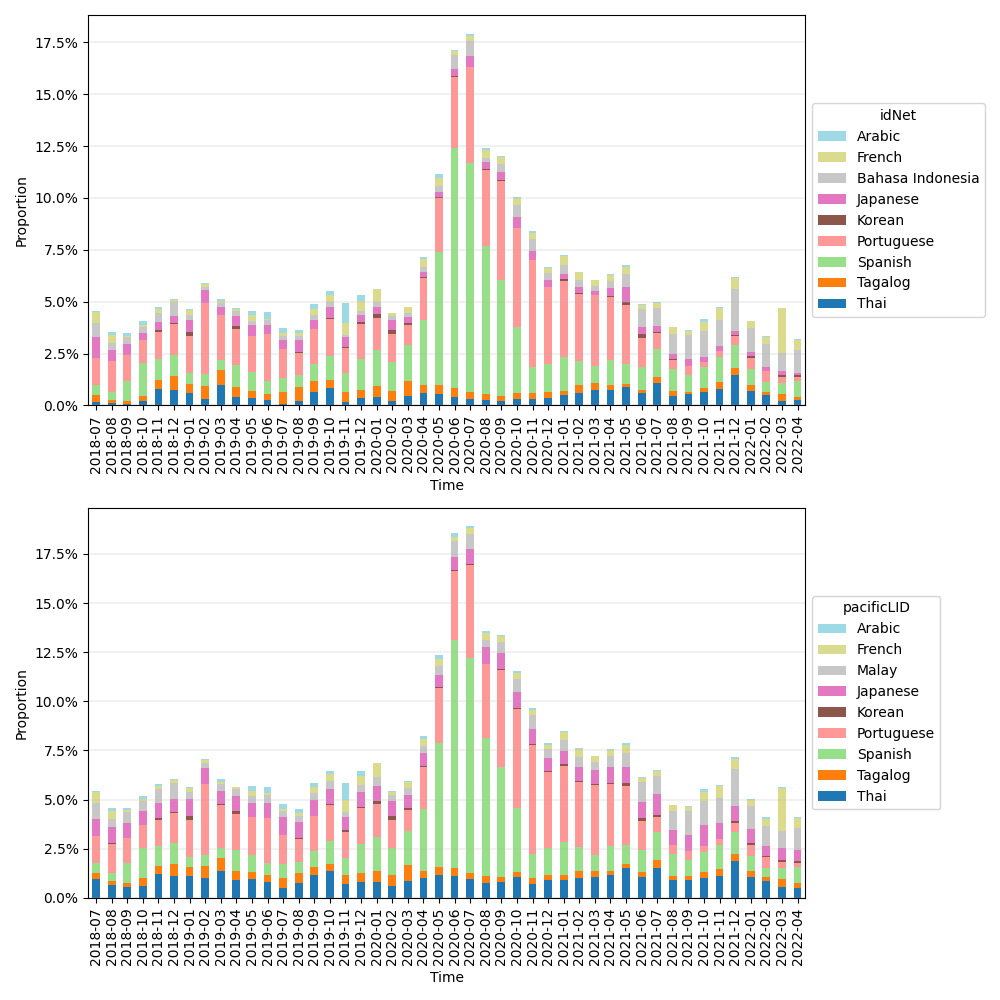}
        \caption{Proportion of languages for the Wellington region by language identification model}
        \label{fig:9}
    \end{figure}
    
    In contrast to the other regions with significant urban areas within New Zealand where linguistic diversity remained stable over times, we observed a significant increase of linguistic diversity in the Wellington region with a peak mid-2020. While other regions also experienced significant fluctuations over time, we could attribute this sampling method where data collection points from rural areas were underrepresented in the social media sub-corpus.
    
    Wellington is the capital region of New Zealand. The Wellington urban area consists of Wellington City, Lower Hutt City, Upper Hutt City, and Pōrirua. The hinterland of the capital region includes the Kāpiti Coast (i.e., Ōtaki) and the Wairarapa (i.e., Masterton, Waipawa). We can see a significant level of overlap in the data catchment area due to the short proximity between the cities within the Wellington urban area. Therefore, we would expect a higher level of internal consistency within the Wellington urban area.
    
    As shown in Figure~\ref{fig:4}, we observed an increase of linguistic diversity in the period between April 2020 and September 2020. A further deep dive of the individual data collection points within the Wellington region revealed a consistent increase of linguistic diversity between the data collection points within the Wellington urban area not observed in the hinterland. We validated this pattern to see if this was a result of data sampling. Figure~\ref{fig:8} provides the monthly $CR$$_{\text{10}}$ measures and mean frequency count of tweets for each data collection point within the Wellington region. We can see the consistent increase of linguistic diversity for Lower Hutt, Pōrirua, Upper Hutt, and Wellington. However, we can see a consistent volume of tweets sampled from each data collection point when we refer to the mean frequency count of tweets. We therefore conclude that this effect is not the result of sampling, but a change in linguistic behaviour within the Wellington urban area.
    
    In Figure~\ref{fig:9}, we provide a stacked bar graph of the most common languages identified by idNet and pacificLID broken down by month for the Wellington region. We removed English from the stacked bar graph to improve interpretability. It is evident that the significant increase of linguistic diversity was due to actual changes in linguistic behaviour. We can observe a first wave increase of tweets in Spanish which was followed by a second wave increase of tweets in Portuguese. This period coincided with the beginning of the national lockdown and border closures as a result of the Covid-19 pandemic. This strongly suggests the Spanish and Portuguese tweets produced by the users were based in New Zealand at the time. This increase is unexpected for the Wellington region as the majority of Spanish and Portuguese speakers in Aoteaora are located in the Auckland region \cite{stats_nz_2020a}.
     
    The coronavirus disease was known colloquially as `corona' before it was officially named `Covid-19' by the World Health Organisation. We considered the possibility that the language identification models erroneously classified tweets with this specific string as Spanish or Portuguese. When we removed these two strings, it did not have an impact on the $CR$$_{\text{10}}$ or proportion of languages for the Wellington region. We also considered the increase of tweets in Spanish and Portuguese was the result of protests and civil unrest across Latin America during this period. Intuitively, this is a reasonable assumption as Wellington is the capital region. The increase of tweets in Spanish and Portuguese during this period remains unresolved and requires further analysis on the content of the tweets themselves which is beyond the scope of the current chapter.

\section{Discussion}
\label{sec:5}

    \paragraph{\textit{How do census data and social media data compare in terms of their basic characteristics and what they might tell us about language use variation over space and time?}}
    
    We acknowledge there are significant conceptual differences between the census data and the social media language data. For example, the purpose of the census data is to collect information on articulated languages (such as spoken and signed) based on self-report, whereas social media data does not include this information. This is evident in the differences in the most common languages used between different spaces as shown in Table~\ref{tab:3} and the differences in measures of linguistic diversity as shown in Table~\ref{tab:4}. 
    
    One unexpected difference in the social media sub-corpus is that none of the Sinitic languages appeared in the top ten list from either models. This is consistent with previous research where user's with written vernacular Chinese as their language background or user interface language were significantly underrepresented in a survey of geotagging enabled users \cite{sloan_2015}. This indicates that the choice of social media platform may lead to differences in how well certain languages are represented.
    
    Although there are differences in the basic characteristics of the two data sources, this does not mean social media language is without its benefits. We can see from our results that linguistic behaviour on social media is sensitive to real-world events in the case of te reo Māori as shown in Figure~\ref{fig:3} and an increase in linguistic diversity in the Wellington region as discussed in Section~\ref{subsec:15}. This otherwise would not be possible with census data.
    
    Frequency counts and linguistic diversity were only two measures we compared between the two data sources; however, social media language data allows us to observe other forms of linguistic behaviour. We can measure the level of code-switching or translanguaging behaviour as we have access to the linguistic signal. It is also possible to see how different language conditions vary over space and time by breaking down the signal into different levels of analysis (e.g., at the word or sentence level).
    
    \paragraph{\textit{How might we use social media data in place of official statistics when performing analyses based on language use information?}}
    
    With reference to the functions of language census topic as discussed in Section~\ref{subsec:2}, social media language data can only indirectly support the needs of culturally and linguistically diverse communities. For example, we could potentially use social media language data to formulate, target and monitor policies and programmes to revitalise te reo Māori or understand the diversity and diversification of New Zealand over time; however, the insights taken from social media may not be a balanced sample of New Zealand's population.
    
    This is because the sample frame and purpose between the census and the social media data are not equivalent. The census provides better coverage of the entire population and spatial granularity, however, it lacks in temporal granularity. This is an advantage of social media data as demonstrated in Figure~\ref{fig:4} and Figure~\ref{fig:5} where we can observe changes in linguistic diversity at a regional geographies. However, we may need to up sample rural regions to ensure our sample is representative of the population.
    
    Another advantage to the social media data is the direct access to linguistic behaviour and how people are using language in New Zealand. Some components of linguistic behaviour we can observe include linguistic content, style, sentiment, and structure. These aspects of linguistic behaviour cannot be observed from a national census or survey. It will be useful to revisit the Wellington case study as discussed in Section~\ref{subsec:15} with additional methods from natural language processing such as topic or sentiment analysis to determine why there was an increase in Spanish and Portuguese during that period.
    
    In a similar vein to administrative data where participant consent is not explicitly given, there are also ethical concerns with social media data \cite{williams_2017}. We need to consider how the use of social media language data for official statistical purposes uphold MDS and the potential risks this may impose on Māori communities across New Zealand \cite{tmr_2018}. We can suggest using alternative data sources alongside official statistics to enrich our understanding of the changing linguistic profile and linguistic behaviour of New Zealand.

\section{Conclusion}
\label{sec:6}

    The results from the current study suggest that we can use online social media language data to observe spatial and temporal changes in linguistic diversity for low-level regional and local geographies. We should be cautious in how we interpret trends and how they can be applied to policy and research as there are conceptual differences between ground truth official statistics and alternative data sources like social media. This does limit the conclusions we can draw from our current analysis as further data validation is required. Despite these limitations, the current chapter provides promising results for alternative data sources to be used alongside census information. Census provides a snapshot of a location at a specific time point, while social media data provides more contemporaneous information about a place. The information available to policy makers and researchers from social media, provides a rich source of language data us to observe real-time changes in linguistic behaviour.

\bibliographystyle{spbasic}
\bibliography{bibliography}

\newpage

\section*{Appendix}
\label{sec:7}

    \begin{table}
        \caption{Data collection points and regional council areas}
        \label{tab:5}
        \begin{tabular}{l p{6cm} c c}
            \hline\noalign{\smallskip}
            Region & Data Collection Points & No. Tweets & \% of Corpus\\
            \noalign{\smallskip}\svhline\noalign{\smallskip}
            Northland & Dargaville, Kawakawa, Kerikeri, Moerewa, Ngunguru, Paihia, Taipa, Waimate North, Whangārei & 372,366 & 3.7\%\\
            Auckland & Auckland, North Shore, Parakai, Waitakere, Warkworth, Wellsford & 1,850,642 & 18.5\%\\
            Waikato & Coromandel, Hamilton, Muriwai Beach, Ngātea, Otorohanga, Paeroa, Pukekohe East, Raglan, Tairua, Taupō, Te Kauwhata, Thames, Tokoroa, Tūrangi, Waihi, Waiuku, Whangamata, Whitianga & 2,133,361 & 21.3\%\\
            Bay of Plenty & Edgecumbe, Katikati, Kawerau, Murupara, Ōpōtiki, Rotorua, Tauranga, Waihi Beach, Whakatāne & 383,597 & 3.8\%\\
            Gisborne & Gisborne & 49,535 & 0.5\%\\
            Hawkes Bay & Hastings, Napier, Wairoa & 218,106 & 2.2\%\\
            Taranaki & Eltham, Hāwera, New Plymouth, Ōpunake, Patea, Waitara & 262,045 & 2.6\%\\
            Manawatū-Wanganui & Bulls, Foxton, Levin, Manakau, Palmerston North, Waiouru, Wanganui & 658,375 & 6.6\%\\
            Wellington & Lower Hutt, Masterton, Ōtaki, Porirua, Upper Hutt, Waipawa, Wellington & 1,244,145 & 12.4\%\\
            West Coast & Greymouth, Hokitika, Westport & 136,344 & 1.4\%\\
            Canterbury & Amberley, Burnham, Christchurch, Darfield, Leeston, Lincoln, Methven, Oxford, Pleasant Point, Rolleston, Timaru, Woodend & 1,379,036 & 13.8\%\\
            Otago & Dunedin, Ōamaru, Queenstown, Wānaka & 446,041 & 4.5\%\\
            Southland & Balclutha, Bluff, Gore, Invercargill, Milton, Riverton, Te Anau, Winton & 237,929 & 2.4\%\\
            Tasman & Brightwater, Māpua, Motueka, Tākaka, Wakefield & 405,008 & 4.0\%\\
            Marlborough & Blenheim, Picton & 235,719 & 2.4\%\\
        \end{tabular}
    \end{table}
    
    \begin{table}
        \caption{Demographic summary of regional council areas}
        \label{tab:6}
        \begin{tabular}{l c c c c }
            \hline\noalign{\smallskip}
            Region & Pop. Density & Median Age\\
            \noalign{\smallskip}\svhline\noalign{\smallskip}
            Northland & 14.3 & 42.6\\
            Auckland & 318.1 & 34.7\\
            Waikato & 19.2 & 37.4\\
            Bay of Plenty & 25.6 & 40.2\\
            Gisborne & 5.7 & 37.0\\
            Hawkes Bay & 11.8 & 40.6\\
            Taranaki & 16.2 & 40.0\\
            Manawatū-Wanganui & 10.7 & 39.4\\
            Wellington & 63.0 & 37.2\\
            West Coast & 5.4 & 46.0\\
            Canterbury & 4.5 & 45.5\\
            Otago & 1.4 & 45.7\\
            Southland & 13.5 & 38.7\\
            Tasman & 7.2 & 38.2\\
            Marlborough & 3.1 & 39.8\\
        \end{tabular}
    \end{table}

\end{document}